\documentclass[letterpaper]{article} 
\usepackage{aaai23}  
\usepackage{times}  
\usepackage{helvet}  
\usepackage{courier}  
\usepackage[hyphens]{url}  
\usepackage{graphicx} 
\usepackage{natbib}  
\usepackage{caption} 
\usepackage{algorithm}
\usepackage{algorithmic}

\usepackage{amsmath}
\usepackage{booktabs}
\usepackage{multirow}
\usepackage{graphicx}
\usepackage{lipsum}
\usepackage{newfloat}
\usepackage{listings}
\DeclareCaptionStyle{ruled}{labelfont=normalfont,labelsep=colon,strut=off} 
\floatstyle{ruled}
\newfloat{listing}{tb}{lst}{}
\floatname{listing}{Listing}
\title{A Multi-View Joint Learning Framework for Embedding Clinical Codes and Text Using Graph Neural Networks}
\author{
    Lecheng Kong,
    Christopher King,
    Bradley Fritz,
    Yixin Chen
}
\affiliations{
    Washington University in St. Louis\\

    One Brookings Drive\\
    St. Louis, Missouri 63130, USA\\
    \{jerry.kong, christopherking, bafritz, ychen25\}@wustl.edu
}

\begin{document}

\maketitle

\begin{abstract}
Learning to represent free text is a core task in many clinical machine learning (ML) applications, as clinical text contains observations and plans not otherwise available for inference.
State-of-the-art methods use large language models developed with immense computational resources and training data; however, applying these models is challenging because of the highly varying syntax and vocabulary in clinical free text.
Structured information such as International Classification of Disease (ICD) codes often succinctly abstracts the most important facts of a clinical encounter and yields good performance, but is often not as available as clinical text in real-world scenarios.
We propose a \textbf{multi-view learning framework} that jointly learns from codes and text to combine the availability and forward-looking nature of text and better performance of ICD codes.
The learned text embeddings can be used as inputs to predictive algorithms independent of the ICD codes during inference.
Our approach uses a Graph Neural Network (GNN) to process ICD codes, and Bi-LSTM to process text.
We apply Deep Canonical Correlation Analysis (DCCA) to enforce the two views to learn a similar representation of each patient.
In experiments using planned surgical procedure text, our model outperforms BERT models fine-tuned to clinical data, and in experiments using diverse text in MIMIC-III, our model is competitive to a fine-tuned BERT at a tiny fraction of its computational effort.

We also find that the multi-view approach is beneficial for stabilizing inferences on codes that were unseen during training, which is a real problem within highly detailed coding systems. We propose a labeling training scheme in which we block part of the training code during DCCA to improve the generalizability of the GNN to unseen codes. In experiments with unseen codes, the proposed scheme consistently achieves superior performance on code inference tasks.
\end{abstract}

\section{Introduction}
An electronic health record (EHR) stores a patient's comprehensive information within a healthcare system. 
It provides rich contexts for evaluating the patient's status and future clinical plans. 
The information in an EHR can be classified as structured or unstructured. 
Over the past decade, ML techniques have been widely applied to uncover patterns behind structured information such as lab results \cite{aiinhealth, deepehr, opportunity}. 
Recently, the surge of deep learning and large-scale pre-trained networks has allowed unstructured data, mainly clinical notes, to be effectively used for learning \cite{clinicalbert, biobert, enhancing}. 
However, most methods focus on either structured or unstructured data \textit{only}. 

A particularly informative type of structured data is the International Classification of Diseases (ICD) codes. 
ICD is an expert-identified hierarchical medical concept ontology used to systematically organize medical concepts into categories and encode valuable domain knowledge about a patient's diseases and procedures.

Because ICD codes are highly specific and unambiguous, ML models that use ICD codes to predict procedure outcomes often yield more accurate results than those do not \cite{code2hospital, code2read}. 
However, \textit{the availability of ICD codes is not always guaranteed}. For example, billing ICD codes are generated after the clinical encounter, meaning that we cannot use the ICD codes to predict post-operative outcomes before the surgery.
A more subtle but crucial drawback of using ICD codes is that there might be \textbf{unseen codes} during inference. 
When a future procedure is associated with a code outside the trained subset, most existing models using procedure codes cannot accurately represent the case.
Shifts in coding practices can also cause data during inference to not overlap the trained set. 

On the other hand, unstructured text data are readily and consistently available. 
Clinical notes are generated as free text and potentially carry a doctor's complete insight about a patient's condition, including possible but not known diagnoses and planned procedures. 
Unfortunately, the clinical text is a challenging natural language source, containing ambiguous abbreviations, input errors, and words and phrases rarely seen in pre-training sources.
It is consequently difficult to train a robust model that predicts surgery outcomes from the large volume of free texts. 
Most current models rely on large-scale pre-trained models \cite{clinicalbert, biobert}. 
Such methods require a considerable corpus of relevant texts to fine-tune, which might not be available at a particular facility.
Hence, models that only consider clinical texts suffer from poor performance and incur huge computation costs.

To overcome the problems of models using only text or codes, we propose to learn from the ICD codes and clinical text in a \textbf{multi-view joint learning framework}. We observe that despite having different formats, the text and code data are complementary and broadly describe the same underlying facts about the patient. This enables each learner (view) to use the other view's representation as a regularization function where less information is present. 
Under our framework, even when one view is missing, the other view can perform inference \textit{independently} and maintain the effective data representation learned from the different perspectives, which allows us to train reliable text models without a vast corpus and computation cost required by other text-only models. 

Specifically, we make the following contributions in this paper.
(1) We propose a multi-view learning framework using Deep Canonical Correlation Analysis (DCCA) for ICD codes and clinical notes.
(2) We propose a novel tree-like structure to encode ICD codes by relational graph and apply Relational Graph Convolution Network (RGCN) to embed ICD codes. 
(3) We use a two-stage Bi-LSTM to encode lengthy clinical texts. 
(4) To solve the unseen code prediction problem, we propose a labeling training scheme in which we simulate unseen node prediction during training. 
Combined with the DCCA optimization process, the training scheme teaches the RGCN to discriminate between unseen and seen codes during inference and achieves better performance than plain RGCN.

\section{Related Works}
\textbf{Deep learning on clinical notes.} Many works focus on applying deep learning to learn representations of clinical texts for downstream tasks. 
Early work \cite{boag} compared the performance of classic NLP methods including bag-of-words \cite{bow}, Word2Vec \cite{w2v}, and Long-Short-Term-Memory (LSTM) \cite{lstm} on clinical prediction tasks. 
These methods solely learn from the training text, but as the clinical texts are very noisy, they either tend to overfit the data or fail to uncover valuable patterns behind the text. 
Inspired by large-scale pre-trained language models such as BERT \cite{bert}, a series of works developed transformer models pre-trained on medical notes, including ClinicalBERT \cite{clinicalbert}, BioBERT \cite{biobert}, and PubBERT \cite{pubbert}. 
These models fine-tune general language models on a large corpus of clinical texts and achieve superior performance. 
Despite the general nature of these models, the fine-tuning portion may not translate well to new settings.
For example, PubBERT is trained on the clinical texts of a single tertiary hospital, and the colloquial terms used and procedures typically performed may not map to different hospitals.
BioBERT is trained on Pubmed abstracts and articles, which also is likely poorly representative of the topics and terms used to, for example, describe a planned surgery.

Some other models propose to use joint learning models to learn from the clinical text, and structured data (e.g., measured blood pressure and procedure codes) \cite{combine2, combining}. Since the structured data are less noisy, these models can produce better and more stable results. However, most assume the co-existence of text and structured data at the inference time, while procedure codes for a patient are frequently incomplete until much later. 

\textbf{Machine learning and procedure codes.} Procedure codes are a handy resource for EHR data mining. Most works focus on automatic coding, using machine learning models to predict a patient's diagnostic codes from clinical notes \cite{towardsbert, icdlstm}. Some other works directly use the billing code to predict clinical outcomes \cite{code2read, code2hospital}, whereas our work focuses on using the high correlation of codes and text data to augment the performance of each. Most of these works exploit the code hierarchies by human-defined logic based on domain knowledge. In contrast, our proposed framework uses GNN and can encode arbitrary relations between codes.    

\textbf{Graph neural networks.} A series of works \cite{gin, mpgnn} summarize GNN structures in which each node iteratively aggregates neighbor nodes' embedding and summarizes information in a neighborhood. The resulting node embeddings can be used to predict downstream tasks. RGCN \cite{rgcn} generalizes GNN to heterogeneous graphs where nodes and edges can have different types. Our model utilizes such heterogeneous properties on our proposed hierarchy graph encoding. Some works \cite{hsgnn, gnnonehr} applied GNN to model interaction between EHRs, whereas our model uses GNN on the code hierarchy.

\textbf{Privileged information.} Our approach is related to the Learning Under Privileged Information (LUPI) \cite{pinfo} paradigm, where the privileged information is only accessible during training (in this case, billing code data). Many works have applied LUPI to other fields like computer vision \cite{deeppinfo} and metric learning \cite{metricpinfo}.

\section{Methods}
\begin{figure}[t]
    \centering
    \includegraphics[width=0.47\textwidth]{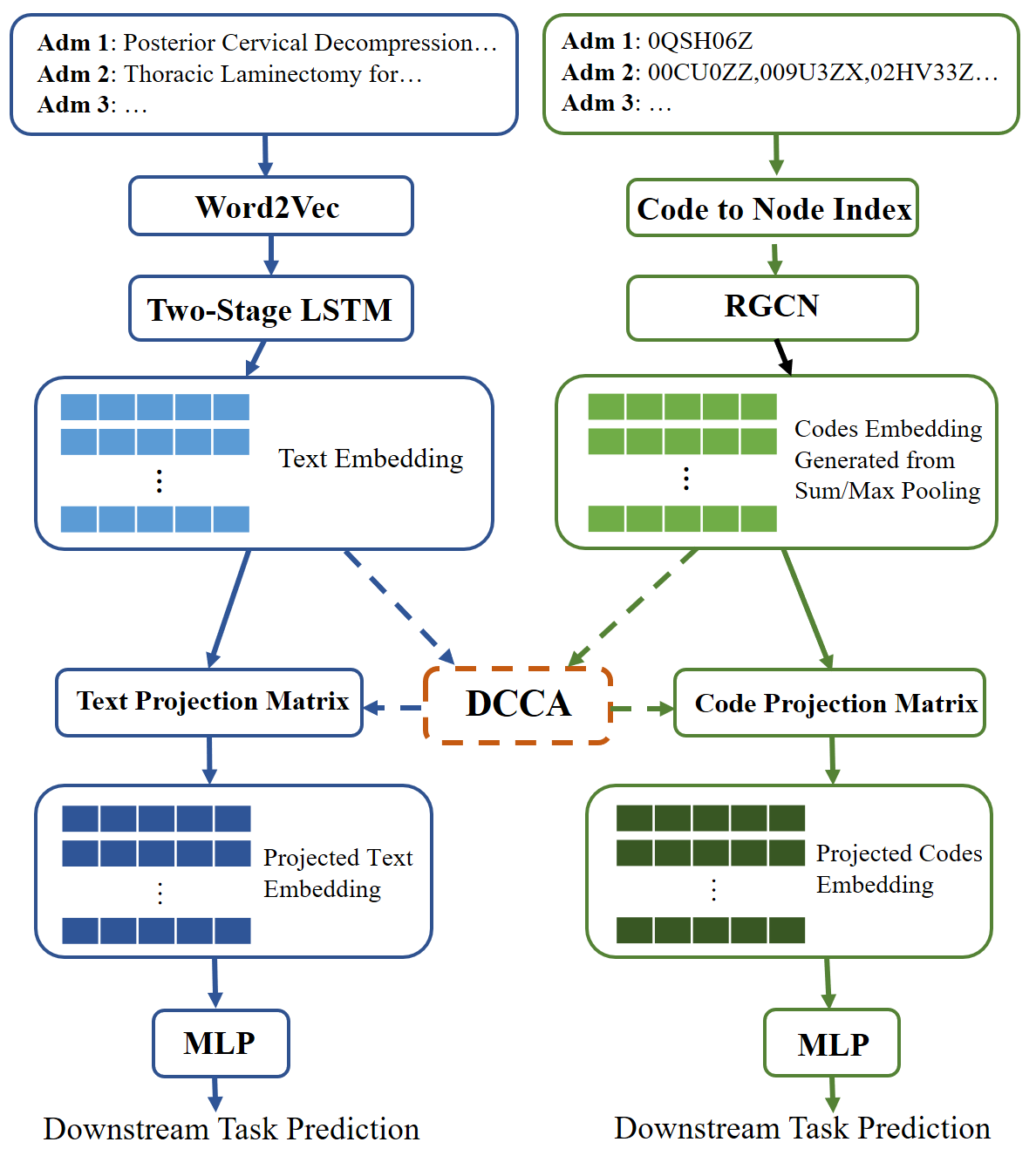}
    \caption{\textbf{Overall multi-view joint learning framework.} Blue boxes/arrows represent the text prediction pipeline, and green represents the code prediction pipeline. Dashed boxes and arrows denote processes only happening during training. By removing the dashed parts, text and code pipelines can predict tasks independently.}
    \label{fig:pipeline}
    \vspace*{-0.2in}
\end{figure}
Admissions with ICD codes and clinical text can be represented as $D=\{(C_1,A_1,y_1),...,(C_n,A_n,y_n)\}$, where $C_i$ is a set of ICD codes for admission $i$, $A_i$ is a set of clinical texts, and $y_i$ is the desired task label (e.g. mortality, re-admission, etc.). The ultimate goal is to minimize task-appropriate losses $\mathcal{L}$ defined as:
\begin{equation} \label{mainloss1}
    \min_{f_C,g_C}\sum_{i}\mathcal{L}(f_{C}(g_{C}(C_i)),y_i)
\end{equation}
and
\begin{equation} \label{mainloss2}
    \min_{f_A,g_A}\sum_{i}\mathcal{L}(f_{A}(g_{A}(A_i)),y_i),
\end{equation}
where $g_{C}$ and $g_{A}$ embed codes and texts to vector representations respectively, and $f_{C}$ and $f_{A}$ map representations to the task labels. 
Note that $(g_{C}, f_{C})$ and $(g_{A}, f_{A})$ should operate independently during inference, meaning that even when one type of data is missing, we can still make accurate predictions.

In this section, we first propose a novel ICD ontology graph encoding method and describe how we use Graph Neural Network (GNN) to parameterize $g_{C}$. We then describe the two-stage Bi-LSTM ($g_{A}$) to embed lengthy clinical texts. We then describe how to use DCCA on the representation from $g_{C}$ and $g_{A}$ to generate representations that are less noisy and more informative, so the downstream models $f_{C}$ and $f_{A}$ are able to make accurate predictions. Figure \ref{fig:pipeline} shows the overall architecture of our multi-view joint learning framework. 

\subsection{ICD Ontology as Graphs}
\begin{figure}
    \centering
    \includegraphics[width=0.47\textwidth]{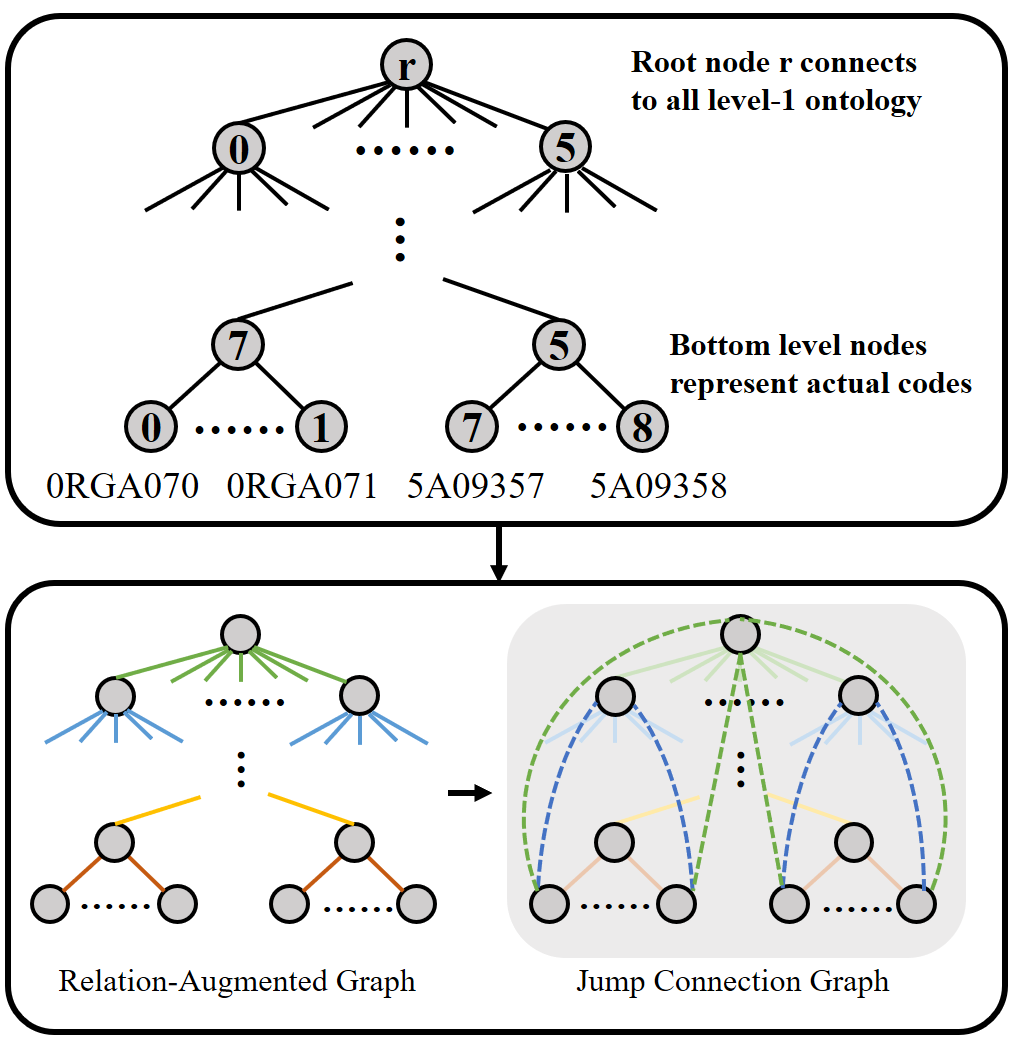}
    \caption{Top: Conventional encoding of ICD ontology. Bottom Left: ICD ontology encoded with relations. Relation types for different levels are denoted by different colors. Bottom Right: Jump connection creates additional edges to leaf nodes' predecessors, denoted by dashed color lines.}
    \label{fig:graph}
    \vspace{-0.3in}
\end{figure}

The ICD ontology has a hierarchical scheme. We can represent it as a tree graph as shown in Figure \ref{fig:graph}, where each node is a medical concept and a node's children are finer divisions of the concept. All top-level nodes are connected to a root node. 
In this tree graph, only the leaf nodes correspond to observable codes in the coding system, all other nodes are the hierarchy of the ontology. 
This representation is widely adopted by many machine learning systems \cite{hap, multimod} as a refinement of the earlier approach of grouping together all codes at the top level of the hierarchy.
A tree graph is ideal for algorithms based on message passing. It allows pooling of information within disjoint groups, and encodes a compact set of neighbors. However, it (1) ignores the granularity of different levels of classification, and (2) cannot encode similarities of nodes that are distant from each other.
This latter point comes about because a tree system may split on factors that are not the most relevant for a given task, such as the same procedure in an arm versus a leg, or because cross-system concepts are empirically very correlated in medical syndromes, such as kidney failure and certain endocrine disorders.

To overcome the aforementioned problems, we propose to augment the tree graph with edge types and jump connections. 
Unlike conventional tree graphs, where all edges have the same edge type, we use different edge types for connections between different levels in the tree graph as shown in the bottom left of Figure \ref{fig:graph}. 
For example, ICD-10 codes have seven characters and hence eight levels in the graph (including the root level). 
The edges between the root node and its children have edge Type 1, and the edges between the seventh level and the last level (actual code level) have edge Type 7. 
Different edge types not only encode whether two procedures are related but also encode the level of similarity between codes.

With multiple edge types introduced to the graph, we are able to further extend the graph structure by jump connections. 
For each leaf node, we add one additional edge between the node and each of its predecessors up to the root node, as shown in the bottom right of Figure \ref{fig:graph}. 
The edge type depends on the level that the predecessor resides. 
For example, in the ICD-10 tree graph, a leaf node will have seven additional connections to its predecessors. Its edge to the root node will have Type 8 (the first seven types are used to represent connections between levels), and its edge to the third level node will have Type 10. 
Jump connections significantly increase the connectivity of the graph. 
Meanwhile, we still maintain the good hierarchical information of the original tree graph because the jump connections are represented by a different set of edge types. 
Using jump connection helps uncover relationships between codes that are not presented in the ontology. 
For example, the relationship between anemia and renal failure can be learned using jump connection even though these diverge at the root node in ICD-9 and ICD-10. 
Moreover, GNNs suffer from over-smoothing, where all node representations converge to the same value when the GNN has too many layers \cite{oversmooth}. 
If we do not employ jump connections, the maximal distance between one leaf node to another is twice the number of levels in the graph. To capture the connection between the nodes, we will need a GNN with that many layers, which is computationally expensive and prone to over-smoothing. 
Jump connections make the distance between two leaf nodes two, and this ensures that the GNN is able to embed any correlation between two nodes. We will discuss this in more detail in Section \ref{sec:gnn}.

\subsection{Embedding ICD Codes using GNN}\label{sec:gnn}
We use GNN to embed medical concepts in the ICD ontology. Let $G=\{V,E,R\}$ be a graph, where $V$ is its set of the vertex (medical concepts in the ICD graph), $E\subseteq\{V\times V\}$ is its set of edges (connects medical concept to its sub-classes), and $R$ is the set of edge type in the graph (edges in different levels and jump connection). As each ICD code corresponds to one node in the graph, we use code and node interchangeably.

We adopt RGCN \cite{rgcn}, which iteratively updates a node's embedding from its neighbor nodes. Specifically, the $k^{th}$ layer of RGCN on node $u\in V$ is: 
\begin{equation}
    h_u^{(k+1)}=\sigma\left(\sum_{r\in R}\sum_{v\in\mathcal{N}_u^r}\frac{1}{c_{u,r}}W_r^{(k)}h_v^{(k)}+W^{(k)}h_u^{(k)}\right)
\end{equation}
where $\mathcal{N}_i^r$ is the set of neighbors of $i$ that connects to $i$ by relation $r$, $h_i^{(k)}$ is the embedding of node $i$ after $k$ GNN layers, $h_i^{0}$ is a randomly initialized trainable embedding, $W_r^{(k)}$ is a linear transformation on embeddings of nodes in $\mathcal{N}_i^r$, $W^{(k)}$ updates the embedding of $u$, and $\sigma$ is a nonlinear activation function. We have $c=|\mathcal{N}_i^r|$ as a normalization factor.

After $T$ iterations, $h_u^{(T)}$ can be used to learn downstream tasks. Since a patient can have a set of codes, $C_i=\{v_{i1},v_{i2},v_{i3},...\}\subseteq V$, we use sum and max pooling to summarize $C_i$ in an embedding function $g_C$:
\begin{equation}
    g_C(C_i)=\sum_{v\in C_i}h_v^{(T)}\oplus \max(\{h_v^{(T)}|v\in C_i\}),
\end{equation}
where $\max$ is the element-wise maximization, and $\oplus$ represents vector concatenation. 
Summation more accurately summarizes the codes' information, while maximization provides regularization and stability in DCCA, which we will discuss in Section \ref{sec:dcca}.

Training RGCN helps embed the ICD codes into vectors based on the defined ontology. Nodes that are close together in the graph will be assigned similar embeddings because of their similar neighborhood. Moreover, distant nodes that appear together frequently in the health record can also be assigned correlated embeddings because the jump connection keeps the maximal distance between two nodes at two. Consider a set of codes $C=\{u,v\}$, because of the summation in the code pooling, using a 2-layer RGCN, we will have non-zero gradients of $h_u^T$ and $h_v^T$ with respect to $h_v^0$ and $h_u^0$, respectively, which connects the embeddings of $u$ and $v$. In contrast, applying RGCN on a graph without jump connections will result in zero gradients when the distance between $u$ and $v$ is greater than two.
               
\subsection{Embedding Clinical Notes using Bi-LSTM}
Patients can have different numbers of clinical texts in each encounter. 
Where applicable, we sort the texts in an encounter in ascending order by time, and have a set of texts $A_i=(a_{i1}, a_{i2},...,a_{in})$. 
In our examples, we concatenate the texts together to a single document $H_i$, and we have $H_i=CAT(A_i)=\bigoplus_{j=\{1...n\}}a_{ij}$.
We leave to future work the possibility of further modeling the collection.

The concatenated text might be very lengthy with over ten thousands word tokens, and RNN suffers from diminishing gradients with LSTM-type modifications. 
While attention mechanisms are effective for arbitrary long-range dependence, they require large sample size and expensive computational resources. Hence, following previously successful approach \cite{clinicalxml}, we adopt a two-stage model which stacks a low-frequency RNN on a local RNN.
Given $H_i$, we first split it into blocks of equal size $b$, $H_i=\{H_{i1},H_{i2},...,H_{iK}\}$. 
The last block $H_{iK}$ is padded to length $b$. 
The two-stage model first generates block-wise text embeddings by
\begin{equation}
    l_{H_{ik}}=LSTM(\{w(H_{ik1}),w(H_{ik2}),...,w(H_{ikb}) \}),
\end{equation}
where $w(\cdot)$ is a Word2Vec \cite{w2v} trainable embedding function. 
The representation of $A_i$ is given by
\begin{equation}
    g_A(A_i)=LSTM(\{l_{H_{i1}},...,l_{H_{iK}}\}).
\end{equation}
The two-stage learning scheme minimizes the effect of diminishing gradients while maintaining the temporal order of the text.

\subsection{DCCA between Graph and Text Data}\label{sec:dcca}
As previously mentioned, ICD codes may not be available at the time when models would be most useful, but are structured and easier to analyze, while the clinical text is readily available but very noisy. 
Despite different data formats, they usually describe the same information: the main diagnoses and treatments for an encounter. 
Borrowing ideas from multi-view learning, we can use them to supplement each other. 
Many existing multi-view learning methods require the presence of both views during inference and are not able to adapt to the applications we envision. 
Specifically, we use DCCA \cite{dcca,deepmv} on $g_A(A_i)$ and $g_C(C_i)$ to learn a joint representation. 
DCCA solves the following optimization problem,
\begin{equation}
    \begin{split}
        \max_{g_C,g_A,U,V}\quad &\frac{1}{N}tr(U^TM_C^TM_AV)\\
        s.t.\quad &U^T(\frac{1}{N}M_C^TM_C+r_CI)U=I,\\
        &V^T(\frac{1}{N}M_A^TM_A+r_AI)V=I,\\
        &u_i^TM_C^TM_Av_j=0,\quad \forall i\neq j, \quad 1\leq i,j\leq L\\
       &M_C=stack\{g_C(C_i)|\forall i\}, \\
       &M_A=stack\{g_A(A_i)|\forall i\},
    \end{split}\label{eq:dcca}
\end{equation}
where $M_C$ and $M_A$ are the matrices stacked by vector representations of codes and texts, $(r_C,r_A)>0$ are regularization parameters. $U=[u_1,...,u_L]$ and $V=[v_1,...,v_L]$ maps GNN and Bi-LSTM output to maximally correlated embedding, and $L$ is a hyper-parameter controlling the number of correlated dimensions. We use $g_C(C_i)U, g_A(A_i)V$ as the final embedding of codes and texts. 
By maximizing their correlation, we force the weak learner (usually the LSTM) to learn a similar representation as the strong learner (usually the GNN) and to filter out inputs unrelated to the structured data. 
Hence, when a record's codes can yield correct results, its text embedding is highly correlated with that of the codes, and the text should also be likely to produce correct predictions.

During development, we found that a batch of ICD data often contains many repeated codes with the same embedding and that a SUM pooling tended to obtain a less than full rank embedding matrix $M_C$ and $M_A$, which causes instability in solving the optimization problem. 
A nonlinear max pooling function helps prevent this.

The above optimization problem suggests full-batch training. However, the computation graph will be too large for the text and code data. Following \cite{deepmv}, we use large mini-batches to train the model, and from the experimental results, they sufficiently represent the overall distribution. After training, we stack $M_C$, $M_A$ again from all data output and obtain $U$ and $V$ as fixed projection matrix from equation \eqref{eq:dcca}.

After obtaining the projection matrices and embedding models, we attach two MLPs ($f_A$ and $f_C$) to the embedding models as the classifier, and train/fine-tune $f_A$ ($f_C$) and $g_A$ ($g_C$) together in an end-to-end fashion with respect to the learning task using the loss functions in \eqref{mainloss1} and \eqref{mainloss2}.

\begin{table*}[h]
\centering
\begin{tabular}{@{}lcccccc@{}}
\toprule
\multirow{2}{*}{Method} & \multicolumn{4}{c}{Local Data} & \multicolumn{2}{c}{MIMIC-III} \\ \cmidrule(l){2-5} \cmidrule(l){6-7}
 & DEL & DIA & TH & D30 & MORT & R30 \\ \midrule
Corr. & $17.3\pm 1.3$ &$16.8\pm 2.6$ & $16.8\pm 2.6$&$16.8\pm 2.6$ &$10.4\pm 1.7$&$12.7\pm 2.3$\\ \midrule
BERT & $65.2\pm 0.6$ & $76.3\pm 1.2$ & $62.1\pm 1.1$ & $74.6\pm 1.8$ & $88.4\pm 1.8$ & $69.2\pm 1.9$ \\
ClinicalBERT & $66.3\pm 0.5$ & $77.0\pm 0.9$ & $\textbf{62.7}\pm 0.8$ & $74.9\pm 1.5$ & $\textbf{90.5}\pm 1.3$ & $\textbf{71.4}\pm 1.8$ \\
Bi-LSTM & $64.6\pm 0.2$ & $76.8\pm 1.8$ & $61.3\pm 1.2$ & $73.9\pm 1.9$ & $87.3\pm 1.7$ & $68.4\pm 2.6$ \\
DCCA+Bi-LSTM & $\textbf{66.9}\pm 0.8$ & $\textbf{78.9}\pm 1.1$ & $61.6\pm 1.1$ & $\textbf{76.5}\pm 1.3$ & $87.2\pm 1.6$ & $71.1\pm 1.4$ \\ \midrule
RGCN & $76.4\pm 1.2$ & $97.2\pm 1.1$ & $75.9\pm 3.0$ & $\textbf{91.5}\pm 1.0$ & $\textbf{90.4}\pm 1.0$ & $\textbf{68.6}\pm 1.4$ \\
DCCA+RGCN & $\textbf{78.9}\pm 1.3$ & $\textbf{98.4}\pm 0.9$  & $\textbf{77.6}\pm 1.2$ & $\textbf{91.5}\pm 1.3$ & $\textbf{90.5}\pm 1.5$ & $67.2\pm 2.5$ \\\midrule
RGCN+Bi-LSTM & $\textbf{79.5}\pm 1.7$ & $97.1\pm 1.4$  & $75.6\pm 0.8$ & $90.8\pm 0.8$ & $\textbf{91.3}\pm 1.2$ & $69.5\pm 1.2$ \\
DCCA+RGCN+Bi-LSTM & $78.7\pm 2.3$ & $\textbf{98.2}\pm 1.3$ & $\textbf{77.1}\pm 2.9$ & $\textbf{91.0}\pm 0.9$ & $90.1\pm 1.3$ & $\textbf{71.2}\pm 1.0$ \\\bottomrule
\end{tabular}
\caption{DCCA Joint Learning and baseline AUROC (\%). Top 4 lines use clinical notes only during inference, middle 2 ICD codes only, and bottom 2 both. Corr = Sum of correlation of latent representations over 20 dimensions.}\label{tab:missv}
\end{table*}
\begin{table*}[h]
\centering
\begin{tabular}{@{}lcccccc@{}}
\toprule
\multirow{2}{*}{Method} & \multicolumn{4}{c}{Local Data} & \multicolumn{2}{c}{MIMIC-III} \\ \cmidrule(l){2-5} \cmidrule(l){6-7}
 & DEL & DIA & TH & D30 & MORT & R30 \\ \midrule
RGCN & $74.6\pm 1.2$ & $87.3\pm 13.1$ & $67.4\pm 6.9$ & $82.8\pm 3.7$ & $84.5\pm 3.6$ & $60.4\pm 2.8$ \\
RGCN+Labling & $73.2\pm 0.6$ & $87.4\pm 14.9$ & $68.5\pm 3.4$ & $83.8\pm 2.1$ & $85.7\pm 3.6$ & $61.3 \pm 2.3$ \\
DCCA+RGCN & $74.9\pm 1.0$ & $89.1\pm 12.5$ & $\textbf{70.8}\pm 0.9$ & $83.5\pm 1.9$ & $85.1\pm 4.1$ & $61.7\pm 2.6$ \\
DCCA+RGCN+Labeling & $\textbf{75.3}\pm 1.1$ & $\textbf{95.4}\pm 0.7$ & $70.6\pm 3.2$ & $\textbf{84.4}\pm 1.4$ & $\textbf{86.4}\pm 4.2$ & $\textbf{63.4}\pm 2.8$ \\ \bottomrule
\end{tabular}
\caption{Ablation Study of the Labeling Training Scheme under Unseen Code Setting in AUROC (\%).}\label{tab:missn}
\end{table*}
\section{Predicting Unseen Codes}
In the previous section, we discuss the formulation of ICD ontology and how we can use DCCA to generate embeddings that share representations across views. 
In this section, we will demonstrate another use case for DCCA-regularized embeddings. 
In real-world settings, the set of codes that researchers observe in training is usually a small subset of the entire ICD ontology. 
In part, this is due to the extreme specificity of some ontologies, with ICD-10-PCS having 87,000 distinct procedures and ICD-10-CM 68,000 diagnostic possibilities before considering that some codes represent a further modification of another entity.
In even large training samples, some codes will likely be seen zero or a small number of times in training.
Traditional models using independent code embedding are expected to function poorly on rare codes and have arbitrary output on previously unseen nodes, even if similar entities are contained in the training data.

Our proposed model and the graph-embedded hierarchy can naturally address the above challenge. 
Its two features enable predictions of novel codes at inference:
\begin{itemize}
    \item \textbf{Relational embedding.} By embedding the novel code in the ontology graph, we are able to exploit the representation of its neighbors. For example, a rare diagnostic procedure's embedding is highly influenced by other procedures that are nearby in the ontology.
    \item \textbf{Jump connection.} While other methods also exploit the proximity defined by the hierarchy, as we suggested above, codes can be highly correlated but remain distant in the graph. Jump connections increase the graph connectivity; hence, our model can seek the whole hierarchy for potential connection to the missing code. Because the connections across different levels are assigned different relation types, our GNN can also differentiate the likelihood of connections across different levels and distances.
\end{itemize}
Meanwhile, during inference, the potential problem is that the model does not automatically differentiate between the novel and the previously seen codes. Because the model never uses novel codes to generate any $g_C(C_i)$, the embeddings of the seen and novel nodes experience different gradient update processes and hence are from different distributions. Nevertheless, during inference, the model will treat them as if they are from the same distribution. However, such transferability and credibility of novel node embeddings are not guaranteed, and applying them homogeneously may result in untrustworthy predictions.


Hence, we propose a labeling training scheme to teach the model how to handle novel nodes during inference. Let $G=\{V,E,R\}$ be the ICD graph and $U$ be the set of unique nodes in the training set, $U\subseteq V$. We select a random subset $U_s$ from $U$ to form the seen nodes during training, and $U_u=V\setminus U_s$ be treated as unseen nodes. We augment the initial node embeddings with 1-0 labels, formally,
\begin{equation}
    \begin{split}
        h_u^{0+}&=h_u^0\oplus 1\quad \forall u\in U_s\\
        h_v^{0+}&=h_v^0\oplus 0\quad \forall v\in V\setminus U_s
    \end{split}
\end{equation} 
Note that we still use $h_u^0$ as the trainable node embedding, while the input to the RGCN is augmented to $h_u^{0+}$. We further extract data that only contain the seen nodes to form the seen data: $D_s=\{(C_i,A_i,y_i)|c\in U_s\forall c\in C_i\}$.

We, again, use DCCA on $D_s$ to maximize the correlation between the text representation and the code representation. After obtaining the projection matrix, we train on the entire dataset $D$ to minimize the prediction loss. Note that $D$ contains nodes that do not appear in the DCCA process and are labeled differently from the seen nodes. The different labels allow the RGCN to tell whether a node is unseen during the DCCA process. If unseen nodes hurt the prediction, it will be reflected in the prediction loss. Intuitively, if unseen nodes are less credible, data with more 0-labeled nodes will have poor prediction results; GNN can capture this characteristic and reflect it in the prediction by assigning less positive/negative scores to queries with more 0-labeled nodes. The labeling training scheme essentially blocks a part of the training code during DCCA and thus obtains embeddings for $U_s$ and $U_u$ from different distributions. And we train on the entire training dataset so that the model learns to handle seen and unseen codes heterogeneously. This setup mimics the actual inference scenario. Note that despite being different, the distributions of seen and unseen node embeddings can be similar and overlapped. Thus, the additional 1-0 labeling is necessary to differentiate them.

\section{Experimental Results}
\textbf{Datasets.} We use two datasets to evaluate the performance of our framework: \textbf{\textit{Proprietary} Dataset.} This dataset contains medical records of 38,551 admissions at the local Hospital from 2018 to 2021. Each entry is also associated with a free text procedural description and a set of ICD-10 \textit{procedure codes}. We aim to use our framework to predict a set of post-operative outcomes, including delirium (DEL), dialysis (DIA), troponin high (TH), and death in 30 days (D30). \textbf{MIMIC-III dataset} \cite{mimic}. This dataset contains medical records of 58,976 unique ICU hospital admission from 38,597 patients at the Beth Israel Deaconess Medical Center between 2001 and 2012. Each admission record is associated with a set of ICD-9 \textit{diagnoses codes} and multiple clinical notes from different sources, including case management, consult, ECG, discharge summary, general nursing, etc. We aim to predict two outcomes from the codes and texts: (1) In-hospital mortality (MORT). We use admissions with hospital\_expire\_flag=1 in the MIMIC-III dataset as the positive data and sample the same number of negative data to form the final dataset. All clinical notes generated on the last day of admission are filtered out to avoid directly mentioning the outcome. We use all clinical notes ordered by time and take the first 2,500-word tokens as the input text. (2) 30-day readmission (R30). We follow \cite{clinicalbert}, label admissions where a patient is readmitted within 30 days as positive, and sample an equal number of negative admissions. Newborn and death admissions are filtered out. We only use clinical notes of type Discharge Summary and take the first 2,500-word tokens as the input text. Sample sizes can be found in Table \ref{tab:sta}.

\textbf{Effectiveness of DCCA training.} We split the dataset with a train/validation/test ratio of 8:1:1 and use 5-fold cross-validation to evaluate our model. GNN and Bi-LSTM are optimized in the DCCA process using the training set. The checkpoint model with the best validation correlation is picked to compute the projection matrix \textit{only from} the training dataset. Then we attach an MLP head to the target prediction model (either the GNN or the Bi-LSTM) and fine-tune the model in an end-to-end fashion to minimize the prediction loss.

For this task, we compare our framework to popular pre-trained models ClinicalBERT and BERT. We also compare it to the base GNN and Bi-LSTM to show the effectiveness of our proposed framework. We additionally provide experimental results where both text and code embedding are used to make predictions. We compare our model with a vanilla multi-view model without DCCA. For all baselines, we report their Area Under Receiver Operating Characteristic (AUROC) as evaluation metrics, and Average Precision (AP) can be found in Appendix \ref{sec:ap}. For all datasets, we set $L$, the number of correlated dimensions to 20, and report the total amount of correlation obtained (Corr).

Table \ref{tab:missv} shows the main results. For clinical notes prediction, we can see that the codes augmented model can consistently outperform the base Bi-LSTM, with an average relative performance increase of 2.4\% on the proprietary data and 1.6\% on the MIMIC-III data. Our proposed method outperforms BERT on most tasks and achieves very competitive performance compared to that of ClinicalBERT. Note that our model only trains on the labeled EHR data without unsupervised training on extra data like BERT and ClinicalBERT do. ClinicalBERT has been previously trained and fine-tuned on the entire MIMIC dataset, including the discharge summaries, and therefore these results may overestimate its performance.

For ICD code prediction, we see that DCCA brings a 1.5\% performance increase on the proprietary data. Since the codes model significantly outperforms the language model on all tasks, the RGCN is a much stronger learner and has less information to learn from the text model. Comparing the results of the proprietary and the MIMIC datasets, we can see that DCCA brings a more significant performance boost to the proprietary dataset, presumably because of the larger amount of correlation obtained in the proprietary dataset (85\% versus 58\%). Moreover, an important difference in these datasets is the ontology used: MIMIC-III uses ICD-9 and the proprietary dataset uses ICD-10. The ICD-9 ontology tree has a height of four, which is much smaller than that of ICD-10 and is more coarsely classified. This may also explain the smaller performance gains in MIMIC-III. 

The combined model with DCCA only brings a slight performance boost compared to the one without because the amount of information for the models to learn is equivalent. Nevertheless, the DCCA model encourages the two views' embeddings to agree and allows independent prediction. In contrast, a vanilla multi-view model does not help the weaker learner learn from the stronger learner.

\textbf{Unseen Codes Experiments.} We identify the set of unique codes in the dataset. We split the codes into k-fold and ran k experiments on each split. For each experiment, we pick one fold as the unseen code set. Data that contain at least one unseen code are used as the evaluation set. The evaluation set is split into two halves as the valid and test sets. The rest of the data forms the training set. We pick another fold from the code split as the DCCA unseen code set. Training set data that do not contain any DCCA unseen code form the DCCA training set. Then, the entire training set is used for task fine-tuning. Because the distribution of codes is not uniform, the number of data for each split is not equal across different folds. We use k=10 for the proprietary dataset and k=20 for the MIMIC-III dataset to generate a reasonable data division. We provide average split sizes in Appendix \ref{sec:unseensize}.

For this task, we compare our method with the base GNN, base GNN augmented with the same labeling training strategy, and DCCA-optimized GNN to demonstrate the outstanding performance of our framework. Similarly, we report AUROC and include AP in Appendix \ref{sec:ap}.

Table \ref{tab:missn} summarizes the results of the unseen codes experiments. Note that all test data contain at least one code that never appears in the training process. In such a more difficult inference scenario, comparing the plain RGCN with the DCCA-augmented RGCN, we see a 2.2\% average relative performance increase on the proprietary dataset. With the labeling learning method, we can further improve the performance gain to 4.2\%. On the MIMIC-III dataset, the performance boost of our model over the plain RGCN is 3.6\%, demonstrating our method's ability to differentiate seen and unseen codes. We also notice that DCCA alone only slightly improves the performance on the MIMIC-III dataset (1.4\%). We suspect that while the labeling training scheme helps distinguish seen and unseen codes, the number of data used in the DCCA process is also reduced. As MORT and R30 datasets are smaller and a small DCCA training set may not faithfully represent the actual data distribution, the regularization effect of DCCA diminishes. 

\begin{table}[t]
\centering
\begin{tabular}{@{}lccc@{}}
\toprule
\multicolumn{1}{r}{} & \# Admission & \# Pos. Samples & \# Unique codes \\ \midrule
DEL & 11,064 & 5,367 & 5,637 \\
DIA & 38,551 & 1,387 & 9,320 \\
TH & 38,551 & 1,235 & 9,320 \\
D30 & 38,551 & 1,444 & 9,320 \\ \midrule
MORT & 5,926 & 2,963 & 4,448 \\
R30 & 10,998 & 5,499 & 3,645 \\ \bottomrule
\end{tabular}
\caption{Statistics of different datasets and tasks.}\label{tab:sta}
\end{table}
\section{Conclusions}
Predicting patient outcomes from EHR data is an essential task in clinical ML. Conventional methods that solely learn from clinical texts suffer from poor performance, and those that learn from codes have limited application in real-world clinical settings. In this paper, we propose a multi-view framework that jointly learns from the clinical notes and ICD codes of EHR data using Bi-LSTM and GNN. We use DCCA to create shared information but maintain each view's independence during inference. This allows accurate prediction using clinical notes when the ICD codes are missing, which is commonly the case in pre-operative analysis. We also propose a label augmentation method for our framework, which allows the GNN model to make effective inferences on codes that are not seen during training, enhancing generalizability. Experiments are conducted on two different datasets. Our methods show consistent effectiveness across tasks. In the future, we plan to incorporate more data types in the EHR and combine them with other multi-view learning methods to make more accurate predictions.
\bibliography{aaai23.bib}
\clearpage

\appendix
\section{Average Precision Score Results}\label{sec:ap}
AP results demonstrate a similar pattern to AUROC results, where DCCA augmented model can consistently outperform the base model while achieving very competitive results compared to ClinicalBERT for the text data as shown in Table \ref{tab:missvap}. The proposed labeling training scheme can also consistently improve our model's performance on the unseen codes experiments, as shown in Table \ref{tab:missnap}.

\section{Hyperparameters}
We use grid search for hyperparameter tuning. For missing view experiments on text, we fix the number of RGCN layers to be 3. We use 32 for all hidden dimensions as we found that varying hidden size has minimal impact on the performance of the data. Text and Code represent the hyperparameters used for text and code inference tasks. Table \ref{tab:hyper} summarizes the set of hyperparameters used for tuning.
\section{Unseen Code Sample Size}\label{sec:unseensize}
We use 10-fold code split for the local data and 20-fold code split for the MIMIC-III data so that the split sizes are reasonable for training. We report the average number of samples for all tasks in Table \ref{tab:unseensize}.
\begin{table}[h!]
\centering
\begin{tabular}{@{}lccc@{}}
\toprule
 & DCCA Train & Full Train & Test \\ \midrule
DEL & 3,458.9 & 4,624.5 & 3,219.4 \\
DIA & 19,305.4 & 23,717.8 & 7,416.3 \\
TH & 19,305.4 & 23,717.8 & 7,416.3 \\
D30 & 19,305.4 & 23,717.8 & 7,416.3 \\
MORT & 4,603.1 & 6,148.2 & 2,424.7 \\
R30 & 2,528.1 & 3,264.0 & 1,330.8 \\ \bottomrule
\end{tabular}
\caption{Average Split Size in Unseen Codes Experiment}\label{tab:unseensize}.
\end{table}
\section{Data And Implementation}
We adopted the local dataset because it is the only dataset we have access to that uses both clinical free texts and ICD-10 codes. The implementation details of the MIMIC-III dataset experiments can be found in the supplementary material (code).
\begin{table*}[htb!]
\centering
\begin{tabular}{@{}lcccccc@{}}
\toprule
 & \multicolumn{4}{c}{Local Data} & \multicolumn{2}{c}{MIMIC-III} \\ \cmidrule(l){2-7} 
 & DEL & DIA & TH & D30 & MORT & R30 \\ \midrule
Corr. & $17.3\pm 1.3$ &$16.8\pm 2.6$ & $16.8\pm 2.6$&$16.8\pm 2.6$ &$10.4\pm 1.7$&$12.7\pm 2.3$\\ \midrule
BERT & $65.4\pm 0.7$ & $23.6\pm 1.2$ & $6.5\pm 1.2$ & $15.1\pm 1.6$ & $85.9\pm 1.9$ & $67.7\pm 1.6$ \\
ClinicalBERT & $\textbf{66.0}\pm 0.6$ & $23.5\pm 0.7$ & $\textbf{7.0}\pm 0.8$ & $15.8\pm 2.1$ & $\textbf{88.6}\pm 1.2$ & $70.1\pm 2.2$ \\
LSTM & $64.4\pm 0.5$ & $22.1\pm 1.6$ & $6.3\pm 0.9$ & $14.3\pm 0.7$ & $85.4\pm 1.4$ & $66.2\pm 2.8$ \\
DCCA+LSTM & $65.4\pm 0.4$ & $\textbf{24.6}\pm 0.8$ & $6.3\pm 1.4$ & $\textbf{15.9}\pm 1.0$ & $84.9\pm 1.6$ & $\textbf{70.4}\pm 2.1$ \\ \midrule
RGCN & $74.2\pm 1.7$ & $\textbf{91.9}\pm 2.1$ & $11.7\pm 0.3$ & $60.4\pm 4.8$ & $90.1\pm 1.7$ & $67.4\pm 2.2$ \\
DCCA+RGCN & $\textbf{76.6}\pm 1.7$ & $90.6\pm 1.6$ & $\textbf{14.8}\pm 0.5$ & $\textbf{60.8}\pm 4.9$ & $\textbf{90.2}\pm 1.2$ & $\textbf{68.4}\pm 1.9$ \\\midrule
RGCN+Bi-LSTM & $\textbf{78.6}\pm 2.6$ & $90.3\pm 1.6$  & $12.6\pm 0.7$ & $\textbf{62.1}\pm 1.5$ & $\textbf{90.2}\pm 1.4$ & $66.8\pm 1.5$ \\
DCCA+RGCN+Bi-LSTM & $77.2\pm 2.1$ & $\textbf{91.6}\pm 2.0$ & $\textbf{15.8}\pm 0.6$ & $61.7\pm 1.2$ & $89.6\pm 1.7$ & $\textbf{67.6}\pm 1.6$ \\ \bottomrule
\end{tabular}
\caption{Effect of DCCA Joint Learning Compared to Different Baselines in AP (\%).}\label{tab:missvap}
\end{table*}
\begin{table*}[htb!]
\centering
\begin{tabular}{@{}lcccccc@{}}
\toprule
\multirow{2}{*}{Method} & \multicolumn{4}{c}{Local Data} & \multicolumn{2}{c}{MIMIC-III} \\ \cmidrule(l){2-7} 
 & DEL & DIA & TH & D30 & MORT & R30 \\ \midrule
RGCN & $72.6\pm 1.5$ & $82.1\pm 9.4$ & $9.2\pm 4.1$ & $53.4\pm 7.1$ & $87.6\pm 3.6$ & $63.7\pm 3.1$ \\
RGCN+Labling & $73.2\pm 0.9$ & $83.6\pm 9.6$ & $9.1\pm 4.7$ & $\textbf{54.2}\pm 9.1$ & $88.5\pm 3.9$ & $\textbf{65.4}\pm 3.6$ \\
DCCA+RGCN & $73.8\pm 1.2$ & $85.3\pm 8.1$ & $12.6\pm 1.3$ & $53.6\pm 8.2$ & $88.7\pm 3.0$ & $65.0\pm 2.9$ \\
DCCA+RGCN+Labeling & $\textbf{74.5}\pm 1.1$ & $\textbf{89.4}\pm 1.3$ & $\textbf{12.7}\pm 3.0$ & $53.5\pm 6.9$ & $\textbf{89.9}\pm 3.1$ & $65.1\pm 3.4$ \\ \bottomrule
\end{tabular}
\caption{Ablation Study of the Labeling Training Scheme under Unseen Code Setting in AP (\%).}\label{tab:missnap}
\end{table*}
\begin{table*}[htb!]
\centering
\begin{tabular}{@{}llcccc@{}}
\toprule
\multicolumn{2}{l}{Hyperparameter} & Local-Text & Local-Code & MIMIC-III-Text & MIMIC-III-Code \\ \midrule
GNN & \#layers & 3 & \{2,3,4\} & 3 & \{2,3,4\} \\ \midrule
LSTM & block size(b) & - & - & 30 & 30 \\ \midrule
\multirow{2}{*}{MLP} & \#layers & 2 & 2 & 1 & 1 \\
 & dropout & \{0,0.2,0.4\} & \{0,0.2,0.4\} & \{0.2,0.4,0.6,0.8\} & \{0.2,0.4,0.6,0.8\} \\ \midrule
\multirow{2}{*}{DCCA} & learning rate & 0.001 & 0.001 & 0.001 & 0.001 \\
 & batch size & 1024 & 1024 & 400 & 400 \\ \midrule
\multirow{2}{*}{Task} & learning rate & \{1e-3,1e-4,1e-5\} & \{1e-3,1e-4,1e-5\} & \{1e-3,1e-4,1e-5\} & \{1e-3,1e-4,1e-5\} \\
 & batch size & 256 & 256 & 32 & 32 \\ \bottomrule
\end{tabular}
\caption{Hyperparameters used for tuning.}\label{tab:hyper}
\end{table*}
\end{document}